\documentclass[preprint,12pt]{elsarticle}

\usepackage{amssymb}
\usepackage{graphicx}
\usepackage{amsthm}
\usepackage{booktabs}
\usepackage{float}

\journal{}

\begin{document}

\begin{frontmatter}



\title{A Dynamic and High-Precision Method for Scenario-Based HRA Synthetic Data Collection in Multi-Agent Collaborative Environments Driven by LLMs}

\author[1]{ {Xingyu Xiao}}
\author[2]{ {Peng Chen}}
\author[1]{{Qianqian Jia}}
\author[1]{{Jiejuan Tong}}
\author[1]{{Jingang Liang}}
\author[1]{{Haitao Wang}}

\affiliation[1]{organization={Institute of Nuclear and New Energy Technology}, organization={Tsinghua University}, city={Beijing}, postcode={100084}, country={China}}       

\affiliation[2]{organization={Software Institute}, organization={Chinese Academy of Sciences},  city={Beijing}, postcode={100086},   country={China}}

\begin{abstract}
HRA (Human Reliability Analysis) data is crucial for advancing HRA methodologies. however, existing data collection methods lack the necessary granularity, and most approaches fail to capture dynamic features. Additionally, many methods require expert knowledge as input, making them time-consuming and labor-intensive. To address these challenges, we propose a new paradigm for the automated collection of HRA data. Our approach focuses on key indicators behind human error, specifically measuring workload in collaborative settings. This study introduces a novel, scenario-driven method for workload estimation, leveraging fine-tuned large language models (LLMs). By training LLMs on real-world operational data from high-temperature gas-cooled reactors (HTGRs), we simulate human behavior and cognitive load in real time across various collaborative scenarios. The method dynamically adapts to changes in operator workload, providing more accurate, flexible, and scalable workload estimates. The results demonstrate that the proposed WELLA (Workload Estimation with LLMs and Agents) outperforms existing commercial LLM-based methods in terms of prediction accuracy.

\end{abstract}

% \begin{highlights}
% \end{highlights}


\end{frontmatter}

\section{Introduction}
\label{Introduction}

Human error is critically important, with statistics indicating that approximately 50\%-80\% of incidents in high-risk industries are caused by human error. Therefore, human reliability analysis (HRA) is a systematic approach to assessing and improving the reliability of human performance in complex systems, particularly in high-risk industries such as nuclear power, aviation, and healthcare. However, most HRA methods rely on expert estimations rather than empirical data \cite{xiao2024krail}. Examples of such methods include human cognitive reliability (HCR) \cite{yu2024human}, standardized plant analysis risk-human reliability analysis (SPAR-H) \cite{gertman2005spar}, and cognitive reliability and error analysis method (CREAM) \cite{hollnagel1998cognitive}. Consequently, numerous scholars have focused on improving data collection in HRA, which is crucial for revolutionizing HRA algorithms. A notable example is the integrated human event analysis system for event and condition assessment (IDHEAS-ECA) \cite{xing2020integrated}, the latest method developed by the NRC based on IDHEAS-DATA \cite{xing2021draft}. Xiao et al. developed a fast and efficient base human error probability solving algorithm based on the IDHEAS-DATA and a large model knowledge graph \cite{xiao2024krail}.

However, HRA data is particularly scarce, and traditional methods typically rely on manually completed surveys to label the data, such as human reliability data extraction (HuREX) \cite{jung2020hurex} and scenario authoring, characterization, and debriefing application (SACADA) \cite{chang2014sacada}, are labor-intensive and time-consuming. Furthermore, small sample sizes of operators for low-probability events present a significant challenge. For example, when dealing with probabilities as low as 1E-3, it may require up to 1,000 data points to observe a naturally occurring error. This often necessitates the introduction of artificial errors in complex scenarios to ensure an adequate dataset for analysis \cite{boring2016goms}. These method is often collected through surveys, which are typically administered static post facto. Boring et al. \cite{boring2023procedure} emphasize the use of the HUNTER dynamic HRA to simulate operator performance. This approach aims to understand both the limitations of synthetic data and its potential advantages compared to expert estimation methods. This method temporarily addresses the issue of static. However, there is still a critical issue, as emphasized by the U.S. NRC: the focus should not solely be on the manifestation of human error but rather on the underlying causes behind human error \cite{xing2020integrated}. The data generated by existing methods predominantly reflect error of omission (EOO) and error of commission (EOC) \cite{xiao2024emergency}. As introduced in Heinrich's Theory and the Iceberg Model in Section \ref{Underlying Factors Contributing to Human Error}, these are merely surface-level manifestations. In order to truly prevent and provide early warnings, we must focus on the underlying indicators behind human error. Therefore, there is a need to develop dynamic methods for generating synthetic underlying metrics behind human error data. 

To address this, a new paradigm for scenario-based HRA data automation collection is proposed, leveraging large language models (LLMs). The scenario we investigate involves multi-agent collaboration, as most real-world tasks require teamwork. Specifically, we use the example of the main control room in a multi-module high-temperature gas-cooled reactor (HTGR) nuclear power plant. In this scenario, the main control room operates under the 'one operator controls two or more reactors' model. The staffing of the multi-module HTGR control room includes three reactor operators (RO1, RO2, RO3), one secondary loop operator (CO), and one shift supervisor (SO).

We begin by collecting workload data through surveys to measure the cognitive load during team collaboration. Building on macro-cognitive theories, we use large models to generate virtual cognitive trajectories. Next, using the Llama-factory framework, we implement the supervised fine-tuning (SFT) process for the Qwen2.5-7B model. Through this fine-tuning, we develop WELLA: Workload Estimation with LLMs and Agents. The results demonstrate that WELLA outperforms current commercial models for predicting workload for roles including RO1, RO2, RO3, CO, and SO.

Section 2 reviews foundational studies on human error, workload estimation models, and the application of large language models for simulating human behavior. Section 3 presents the comprehensive framework employed in this research. Section 4 outlines the experimental design and key findings. Section 5 summarizes the contributions of this research, reflects on its limitations, and proposes directions for future work.

\section{Related Work }

This section presents an in-depth analysis of the underlying factors contributing to human error, explores various workload estimation models, and discusses the application of large language models in simulating human behavior.

\subsection{Underlying Factors Contributing to Human Error} \label{Underlying Factors Contributing to Human Error}
Human error, a critical element in safety analysis across various domains, is often the result of multiple interacting factors. These factors can be broadly categorized into cognitive, environmental, organizational, and physiological influences \cite{boring2007measure}, all of which shape an individual's performance and decision-making process.  Understanding these underlying causes is essential for mitigating risks and improving system design. 

It is often compared to the iceberg theory \cite{chamon2013iceberg}, suggesting that there is much more beneath the surface of visible behavior. Prior to the occurrence of an incident, there are numerous underlying representations that influence the final outcome. The theory posits that an individual's "self" resembles an iceberg, where only a small portion—behavior—is observable above the surface, while the larger, more complex inner world remains hidden beneath. This hidden world encompasses seven layers: behavior, coping mechanisms, emotions, perspectives, expectations, desires, and the self \cite{wright2000towards}. Each of these deeper layers plays a pivotal role in shaping behavior and ultimately influencing the occurrence of errors or accidents.

 Therefore, it is crucial to focus not only on the visible actions during the event but also on the factors and conditions that precede it. 

The Heinrich Law,\cite{domurath2015stress} formulated by the renowned American safety engineer H.W. Heinrich, is a principle commonly known as the 300:29:1 ratio. Based on his analysis of workplace injury and accident statistics, Heinrich proposed this law to provide insights into accident prediction and risk management, particularly for insurance companies. The law stipulates that for every 300 unsafe acts or conditions present in a workplace, 29 will result in minor injuries or incidents, and of these 29 minor incidents, one will inevitably lead to a serious injury, fatality, or major disaster.

This ratio emphasizes the critical importance of addressing hazards before they escalate into accidents. Heinrich's research underscores the need for proactive safety measures and the systematic identification and elimination of risks to prevent catastrophic outcomes. The principle serves as a foundational concept in safety engineering and risk management, influencing policies and safety protocols in various industries.

To prevent human-factor accidents, it is essential not only to focus on external errors but also to pay attention to underlying indicators. Workload serves as a valuable indicator in this regard and is widely applied in high-risk industries such as aviation \cite{kantowitz2017human} and nuclear power plants \cite{jou2009evaluation}. This paper applies the macrocognitive theory to analyze workload from five perspectives—detection, understanding, decision-making, action execution, and inter-team coordination—in order to construct a cognitive trajectory.

\subsection{Workload Estimation Models}\label{Workload Estimation Models}

Workload estimation is a critical component in ensuring optimal human performance, particularly in high-risk industries such as nuclear power, aerospace, and healthcare. A variety of models have been developed to assess the cognitive, physical, and emotional demands placed on operators, with the aim of predicting potential overload and improving safety and efficiency. 

Specifically,workload estimation can be divided into contact-based, non-contact-based 
and a mixed-based measurements. The most common contact-based approach involves EEG-based methods \cite{bagheri2020eeg,brouwer2012estimating}, where EEG signals are collected and analyzed using various deep learning techniques such as CNN, LSTM, and Transformer for prediction and early warning \cite{hassan2024eeg}. In addition, other physiological signals, such as Heart Rate Variability \cite{delliaux2019mental} and eye activity features \cite{marquart2015review}, have also been explored. Some studies combine multiple signals for multimodal inference. For instance, Planke et al. \cite{planke2021online} proposed different configurations to fuse data from an Electroencephalogram (EEG) model’s output, four eye activity features, and a control input feature. Xing et al. \cite{xing2018driver} utilized a variety of sensors to provide signals that can assist in workload estimation. However, measuring EEG, eye movement, and other signals under normal working conditions is not always feasible, as such measurements may interfere with the regular tasks and operations.

For non-contact measurements, the primary approach involves Subjective Measures of Mental Workload \cite{schvaneveldt1998modeling}, such as SWAT, WP, NASA-TLX, RSME, and DALI questionnaires. However, subjective methods can only be used for post-task evaluation, making them a static and passive approach. Despite this limitation, several studies confirm the value of subjective measures. Mental workload is a multi-faceted phenomenon, as reflected in the literature. It can be related to physiological states of stress and effort, subjective experiences of stress, mental effort, and time pressure, as well as objective measures of performance levels and performance breakdowns. Therefore, despite various physiological and psychological assessments, subjective measures indicate that subjects tend to be consistent in their workload ratings. Mental workload is typically categorized into time-load, mental effort load, and psychological stress load.

In addition, there are computer modeling methods for non-contact-based measurements, such as a cognitive architecture for modeling cognition (ACT-R) \cite{ritter2019act} and queueing network-model human processor (QN-MHP) \cite{liu2006queueing}, which calculate workload from a physiological perspective. However, these methods tend to be rigid, requiring pre-set Lisp models or task sequences for each task. These approaches have several drawbacks:
\begin{itemize}
    \item High entry barrier: Users must learn Lisp modeling language and understand the underlying mechanics of the model.
     \item High workload: Different models must be developed for each task, and experiments are needed to validate the accuracy of each model.
     \item Lack of generalization: These methods lack the ability to adapt to novel procedures they have never encountered.
     \item Subjectivity in design: The design of Lisp models and task sequences is subjective, making it difficult to quantitatively explain their results.
\end{itemize}

Many studies combine objective and subjective indicators. For example, Julie Paxion et al. examined how both subjective and objective levels of mental workload influence performance, as a function of situational complexity and driving experience \cite{paxion2014mental}.

Additionally, some studies focus primarily on the physiological perspective, particularly neuroscience. For instance, Luca Longo's data analysis strongly suggests that usability and mental workload are two non-overlapping constructs, and they can be jointly employed to significantly improve the prediction of human performance \cite{longo2018experienced}. Frédéric Dehais et al. conducted research from a neurophysiological perspective \cite{dehais2020neuroergonomics}, though this is outside the scope of the present work.

Overall, there is currently a lack of a novel experimental paradigm that can be non-contact, relatively accurate, and dynamic, while also adapting to task-specific contexts for modeling purposes.

\subsection{Large Language Models in Simulating Human Behavior} \label{Large Language Models}

The emergence of large language models (LLMs) has prompted significant research across various specialized domains. Many scholars have explored the use of LLMs for simulating human behavior. For instance, Wang et al. propose the Learning through Communication (LTC) framework to enhance the human-likeness of agents \cite{wang2024adapting}. Xie et al. demonstrate that LLM agents, especially GPT-4, exhibit strong behavioral alignment with humans in terms of trust behaviors, suggesting the potential of using LLM agents to simulate human trust dynamics \cite{xie2024can}. Additionally, Karthik Sreedhar et al. find that multi-agent systems outperform single LLMs in simulating human reasoning and actions, with an accuracy rate of 88\% compared to 50\% for single agents when applied to personality pair simulations \cite{sreedhar2025simulating}. Wu et al. \cite{wu2024shall} demonstrate that Large Language Models (LLMs) have become progressively prevalent in the domain of social simulations. Their study investigates spontaneous cooperation in the context of three competitive scenarios, successfully replicating the gradual emergence of cooperative behaviors. The results of their simulations exhibit a striking alignment with empirical human behavioral data, providing insights into the dynamics of cooperation in competitive environments.

However, vanilla LLMs do not necessarily provide accurate representations of human behavior. Lindia Tjuatja et al. \cite{tjuatja2024llms} evaluate nine different models and find that popular open-source and commercial LLMs generally fail to exhibit human-like behaviors, particularly in models that have undergone reinforcement learning from human feedback (RLHF). Moreover, even when a model demonstrates a significant alignment with human behavior in certain directions, it remains highly sensitive to perturbations that do not cause substantial changes in human responses. These findings highlight the challenges of using LLMs as proxies for human behavior and emphasize the need for more nuanced characterizations of model behavior. In contrast, supervised fine-tuning (SFT) offers a more promising approach. This process typically involves fine-tuning a pre-trained model through supervised learning, utilizing annotated data to optimize its performance \cite{xiao2024text}. In this paper, we leverage real human data to optimize the model and design multiple agents within a simulated environment to interact, thereby exploring a more effective modeling approach that better represents human behavior.

\section{Methodology}

\subsection{Framework of the study}

\begin{figure}[h]
\centering
\includegraphics[width=1.0\textwidth]{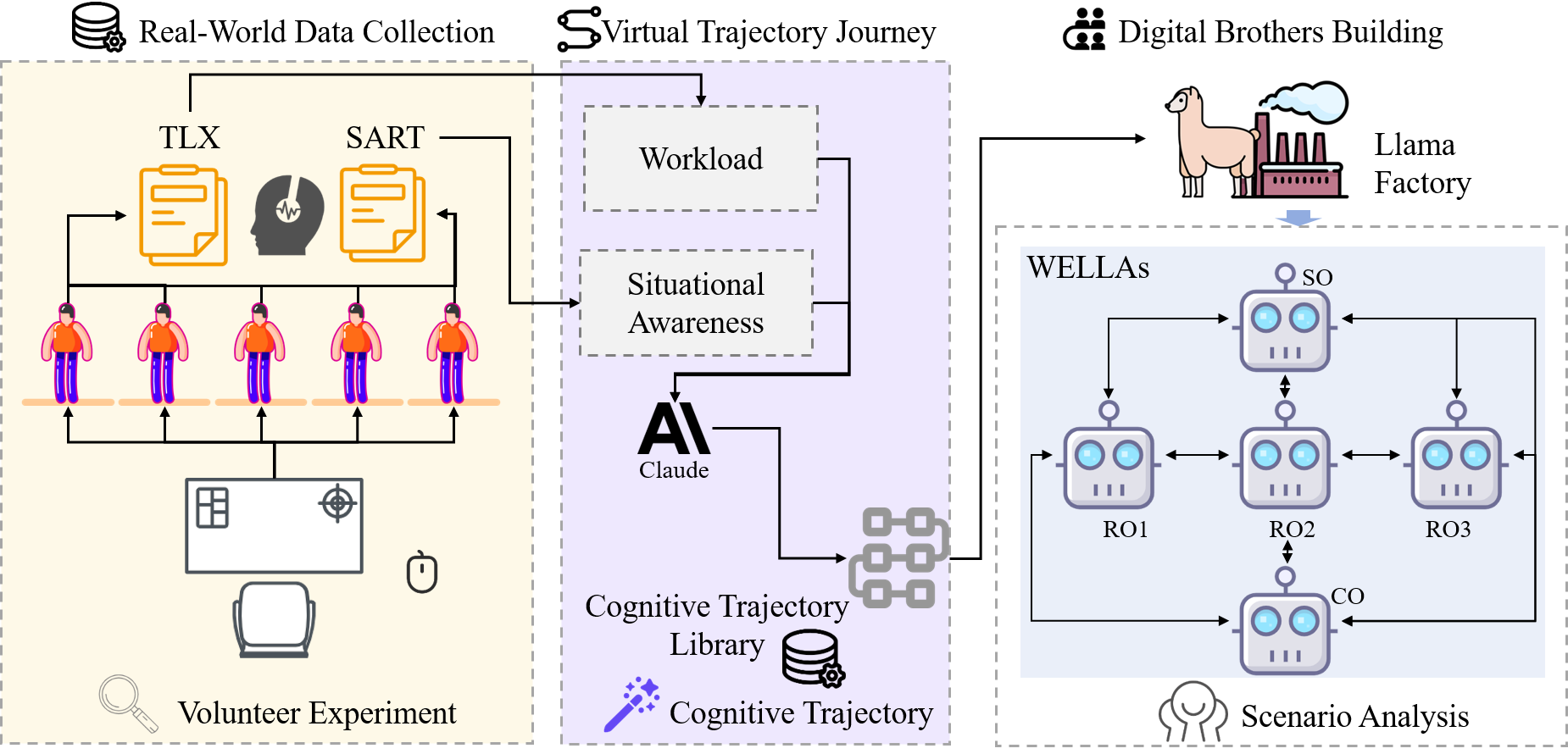}
\caption{IDHEAS-ECA HRA Process }\label{workflow}
\end{figure}

As illustrated in Figure \ref{workflow},  our framework is divided into three key components: real-world data collection, virtual cognitive trajectory generation, and digital brother building. First, we employed a testbed to identify real nuclear power plant operators and collected data from multiple scenarios to train the model's generalization capabilities across different tasks. In the second part, we aim for the proposed framework to simulate human cognitive processes. To achieve this, we utilized Claude to generate the virtual cognitive trajectory. The third part involves training our digital brother for workload estimation. Using the Llama-factory framework, we performed supervised fine-tuning (SFT) on our cognitive trajectory library data. We then constructed five agents tailored to the scenario: SO, RO1, RO2, RO3, and CO, in order to enable multi-agent collaboration for multi-scale workload estimation. The scenario studied in this research involves multi-agent collaboration, as most tasks in practice require collaborative efforts among multiple individuals. The subsequent sections of this paper use the fourth-generation High-Temperature Gas-Cooled Reactor (HTGR) \cite{zhang2006design} as a case study to demonstrate the entire process of the proposed methodology.

\subsection{Real-World Data Collection}

Next, we examine the main control room of a multi-module High-Temperature Gas-Cooled Reactor (HTGR) nuclear power plant. In this setting, the main control room operates under the one operator controls two or more reactors. The control room is staffed by three reactor operators (RO1, RO2, RO3), one secondary loop operator (CO), and one shift supervisor (SO). RO1 is responsible for the operation of the 1 and 2 Nuclear Steam Supply System (NSSS) modules, RO2 oversees the operation of the 3 and 4 NSSS modules, and RO3 manages the operation of the 5 and 6 NSSS modules. The CO is in charge of the conventional island operations, while the SO coordinates the overall activities of the control room.

\begin{figure}[h]
\centering
\includegraphics[width=1.0\textwidth]{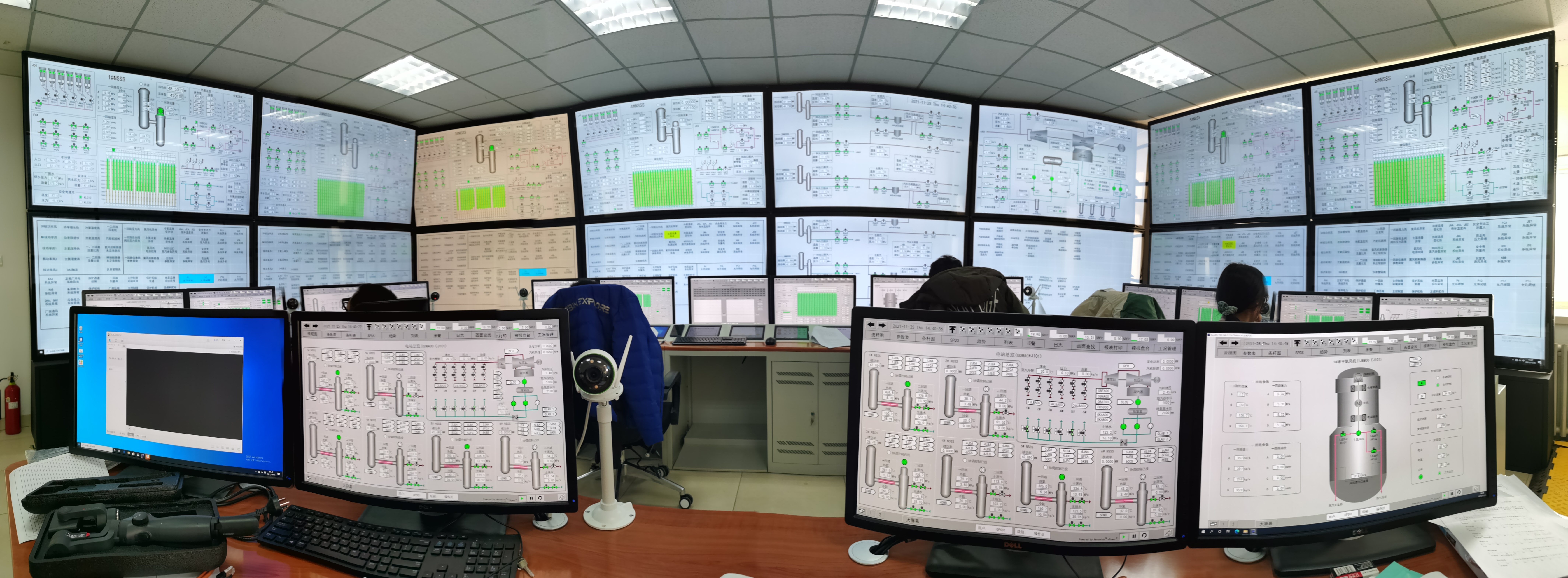}
\caption{The High-Temperature Reactor Simulator Operating Test Bench }\label{simulator}
\end{figure}

Figure \ref{simulator} illustrates the experimental setup used to collect data. We enlisted real operators (RO1, RO2, RO3, SO, CO) to simulate responses under various scenarios. After each simulation, operators were asked to complete the NASA Task Load Index (TLX) \cite{hart1986nasa} and situational awareness measurements based on the Situational Awareness Rating Technique (SART) \cite{taylor2017situational} from real HTGR operators, which were then used to gather performance and workload data.

The NASA-TLX is one of the most established subjective workload measurement tools, designed to assess the perceived workload in task performance across six dimensions: mental demand, physical demand, temporal demand, performance, effort, and frustration  \cite{hart2006nasa}. Each dimension is evaluated by the operator on a scale of 0 to 100, and these scores are then weighted according to the importance of each factor in the context of the task. NASA-TLX has been widely applied in numerous high-risk settings, including aviation, military operations, and healthcare, where accurate workload estimation is essential for managing operator performance and minimizing error. The standard NASA-TLX is typically based on weighted scores, where users are required to assign weightings to each dimension. In this study, an unweighted version of the NASA-TLX method is used for workload calculation equation in equation \ref{TLX-eq}.
\begin{equation}
    workload=\frac{(MD+PD+TD+E+(100-P)+F)}{6} 
\end{equation}\label{TLX-eq}

Where MD represents mental demand, PD represents physical demand, TD represents temporal demand, E stands for effort, P denotes performance, and F indicates frustration.

Furthermore, as we propose a scenario-driven dynamic approach, we also aim to collect situational awareness data to further infer that risk perception is crucial in evaluating an individual’s workload. Therefore, we use the SART scale for this purpose. The SART scale consists of three dimensions: Demand, Supply, and Understand, as detailed in Appendix D. The Demand dimension includes the first three items of the scale, Supply includes items four through seven, and Understand includes items eight through ten. Each item is scored on a range from 1 to 7, and the score for each dimension is the sum of the scores of the items it contains. The calculation method for the operator's Situational Awareness (SA) is given in Equation \ref{SA}

\begin{equation}
    SA=Understand-(Demand -Supply)
\end{equation}\label{SA}

We conducted experimental validation across multiple types of scenarios. The specific experimental scenarios include Startup, Shutdown, and Accident, with the number of instances for each sub-scenario, which are 28, 11, and 30, respectively.

\subsection{Vitural cognitive trajectory Generation}

\begin{figure}[h]
\centering
\includegraphics[width=1.0\textwidth]{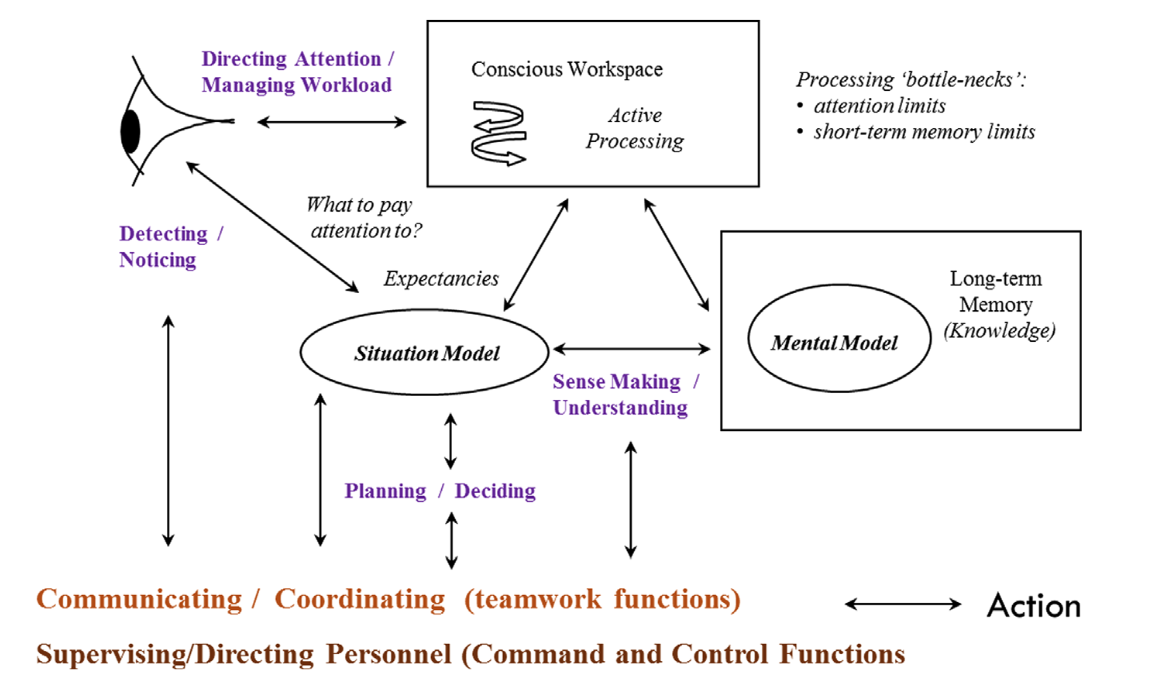}
\caption{A simplified model of macrocognitive functions. \cite{chang2014sacada}}\label{marcro}
\end{figure}

We refer to the macro-cognitive theory \cite{strydom2011cognitive} , which is primarily divided into five categories: detection, understanding, decision-making, action execution, and inter-team coordination. Although it is challenging to obtain real human cognitive trajectorys, studies suggest that a cognitive trajectory indeed exists during human actions \cite{xiao2024krail}, and this is the underlying reason behind workload. Therefore, we leverage large language models (LLMs) to generate human-like cognitive trajectorys, enabling the model to think more like a human. A simplified model of macrocognitive functions is shown in Figure \ref{marcro}

% 训练数据有2846条，
%测试数据有383 条，

\subsection{Language Model Fine-Tuning}

Supervised fine-tuning (SFT) \cite{dong2023abilities} enables large language models (LLMs) to perform specific tasks, such as question answering, dialogue, and reasoning, as anticipated \cite{xiao2024text}. With the availability of instructional data, SFT is utilized to guide the models in executing tasks related to the semiconductor industry. 

We use Qwen2.5-7B \cite{yang2024qwen2} as the base model and fine-tune it using the Llama-factory framework. Qwen2.5-7B is the latest publicly available version, suitable for a wide range of applications, including but not limited to natural language processing, text generation, code writing, and mathematical computations. With its extensive knowledge base and robust encoding and mathematical capabilities, the model is particularly well-suited for tasks requiring these skills. Additionally, it can generate long texts exceeding 8K tokens, understand structured data (e.g., tables), and produce structured outputs, especially in JSON format. The model also supports up to 29 languages, including Chinese, English, French, and Spanish, making it highly effective for multilingual applications. Llama-factory \cite{zheng2024llamafactory}, as the latest Llama framework, is known for its efficiency, significantly reducing training time and computational resources, while offering exceptional flexibility.

% \subsection{Multi-Agent Collaboration}

% \begin{figure}[h]
% \centering
% \includegraphics[width=0.7\textwidth]{pics/muti-agent.png}
% \caption{The High-Temperature Reactor Simulator Operating Test Bench }\label{simulator}
% \end{figure}

\subsection{Scenario Analysis}

\begin{figure}[H]
\centering
\includegraphics[width=0.8\textwidth]{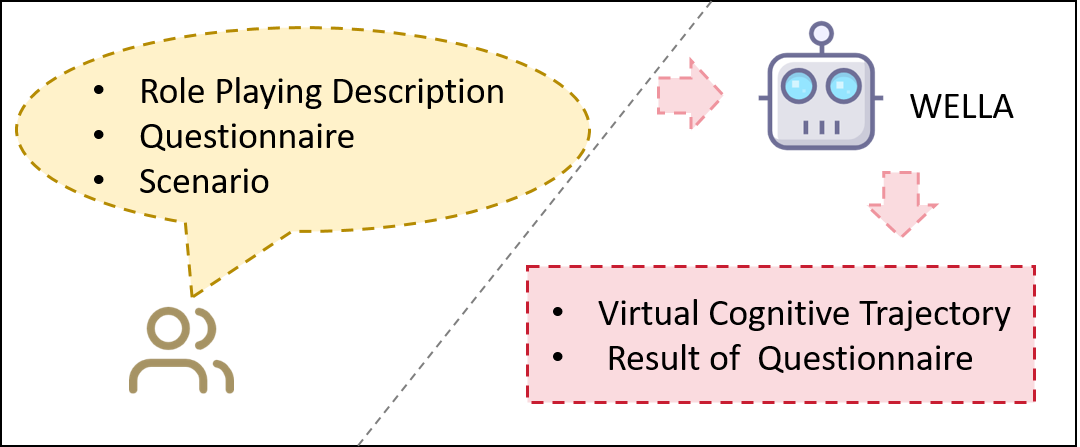}
\caption{WELLA: Input-Output Dynamics in Scenario Analysis }\label{infer}
\end{figure}

The specific generation process for scenario analysis is detailed in Figure \ref{infer}. Given an input scenario, the output consists of the workload for different workers with distinct roles. This process effectively enables the generation of synthetic human workload data.

\section{Experiments and Results}

This section provides a detailed of the implementation details, workload estimation accuracy, comparisons with existing methods, and scenario analysis.

\subsection{Implementation Detail}

For training WELLA, we utilized 2 NVIDIA A800 80GB GPUs. We followed guidelines from transformers huggingFace, HuggingFace Accelerator, and the LLaMA-Factory library for fine-tuning a LLM. The hyperparameters for pretraining and SFT included a batch size of 2 and a learning rate of 1.0e-5. The training was conducted 8 epochs. It is worth noting that , we incorporated SFT data along with a special token \cite{liu2023hierarchical}. The use of special tokens is a validated practice, which helps balance the pretraining and SFT processes of the model. Detailed implementation can be found in \ref{liu2023hierarchical}, and will not be discussed further in this paper.

% for teacher model, train data is 1213

\subsection{Workload Estimation}

\begin{figure}[h]
\centering
\includegraphics[width=1.0\textwidth]{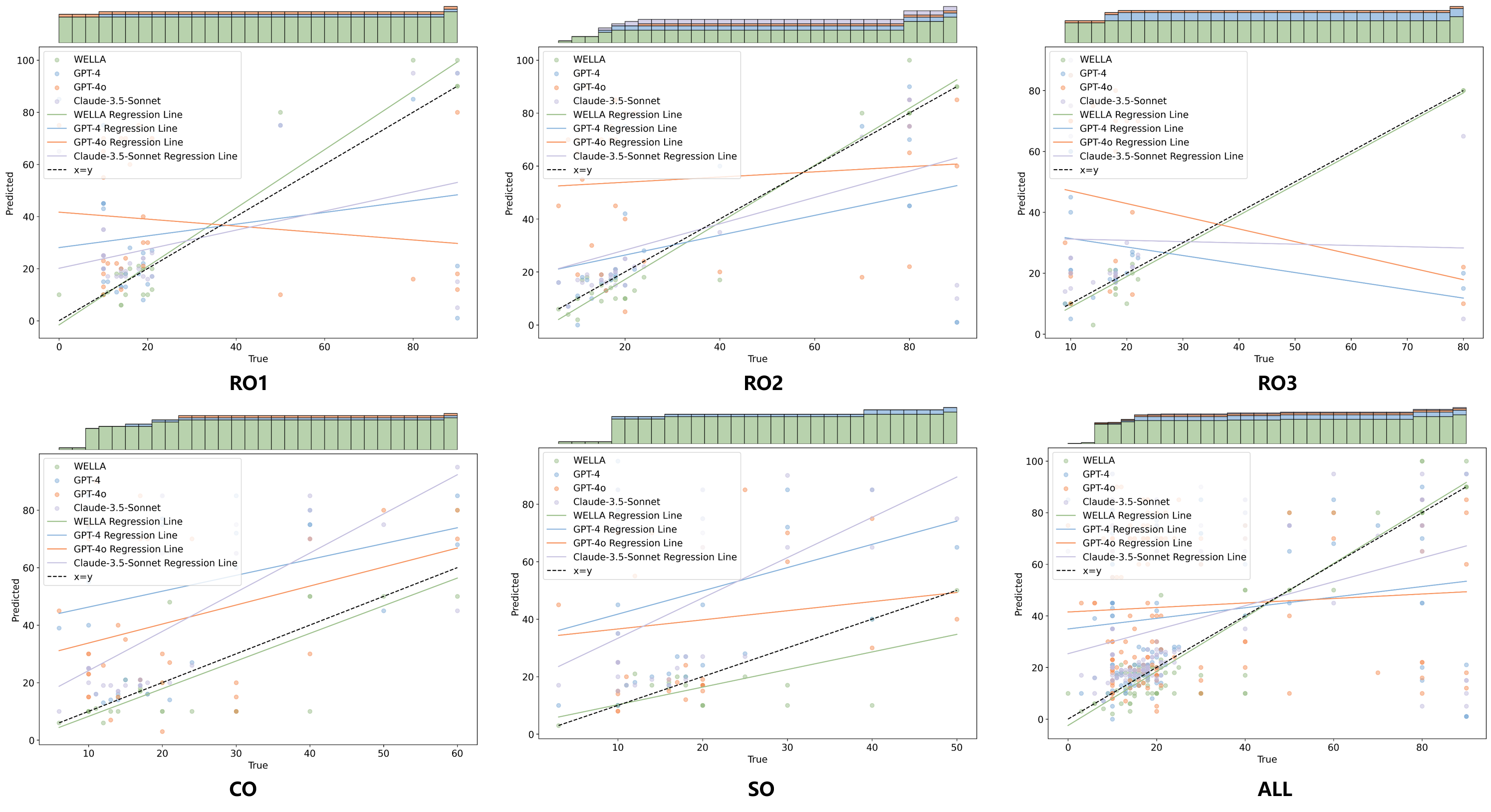}
\caption{Performance Comparison of WELLA, GPT-4, GPT-4O, and Claude Across Multiple Scenarios}\label{results}
\end{figure}

This section presents the results of our model in terms of workload estimation accuracy. To highlight the effectiveness of our model, we compare the output of our WELLA model with the performance of currently unavailable commercial models, including GPT-4 \cite{achiam2023gpt}, GPT-4o \cite{shahriar2024putting}, and Claude-3.5-Sonnet \cite{xiao2024hybrid}. Figure \ref{results} presents a comparative analysis of predictive performance and their alignment with ground truth values across these models. The evaluation is conducted for five scenarios: RO1, RO2, RO3, CO, SO, and the aggregated dataset (ALL).

Regarding role-playing tasks RO1, RO2, RO3, and CO, WELLA's regression line appears closest to the diagonal, indicating predictions that are more consistent with the true values. In contrast, GPT-4, GPT-4o, and Claude-3.5-Sonnet show less alignment with the diagonal, suggesting greater deviations from accurate predictions. As for SO, the model's performance may have been compromised due to the unclear description of SO's responsibilities in this study, where only the input "SO coordinates the overall activities of the control room" was provided. Consequently, the model's predictions for SO are not as robust. However, overall, as shown in Figure \ref{results}, the performance through all data remains relatively the best.

\begin{table}[h]
\caption{ Performance Comparison of Models in Scenario RO1}\label{RO1-table}%
\begin{tabular}{p{4cm}p{2cm}p{2cm}p{2cm}p{2cm}}
\toprule
 & $R^{2} $  & RMSE  & MAE & EV \\
\midrule
GPT-4& -0.7107 & 33.9131 &21.3000 & -0.5911\\
GPT-4o& -1.4378 &40.4837 &31.4667 & -1.1573\\
Claude-3.5-Sonnet& -0.0308& 26.3255&15.8333 & -0.0007\\
WELLA& 0.9012 &8.1507&4.7000 & 0.9040\\

\bottomrule
\end{tabular}
\end{table}

Table \ref{RO1-table} provides a comparative analysis of model performance in Scenario RO1 across four evaluation metrics: coefficient of determination ($R^{2} $), root mean square error (RMSE), mean absolute error (MAE), and explained variance (EV). Among the models, WELLA demonstrates superior performance with a positive $R^{2} $ of 0.9012, the lowest RMSE (8.1507) and MAE (4.7000), as well as the highest EV (0.9040), indicating strong predictive accuracy and variance explanation. In contrast, GPT-4, GPT-4o, and Claude-3.5-Sonnet exhibit negative 
$R^{2} $ and higher error metrics, suggesting weaker predictive capabilities and less effective data representation in this scenario. These results highlight WELLA as the most effective model for Scenario RO1.

\begin{table}[h]
\caption{ Performance Comparison of Models in Scenario RO2}\label{RO2-table}%
\begin{tabular}{p{4cm}p{2cm}p{2cm}p{2cm}p{2cm}}
\toprule
 & $R^{2} $  & RMSE  & MAE & EV \\
\midrule
GPT-4& -0.0825 & 30.2953 &16.5806 & -0.0783\\
GPT-4o& -1.1799 &42.9913 &35.4193 & -0.6090\\
Claude-3.5-Sonnet& 0.0347& 28.6086&13.2903 & 0.0378\\
WELLA& 0.9343& 7.4639 &4.7419& 0.9383\\

\bottomrule
\end{tabular}
\end{table}

Table \ref{RO2-table} presents the performance comparison of models in Scenario RO2. WELLA significantly outperforms other models across all metrics, achieving the highest $R^{2}$ (0.9343) and EV (0.9383), along with the lowest RMSE (7.4639) and MAE (4.7419), indicating its superior predictive accuracy and ability to explain data variability. In contrast, GPT-4 and GPT-4o show negative $R^{2} $ and EV values, along with higher error metrics, reflecting weaker performance. Claude-3.5-Sonnet performs slightly better than GPT-4 and GPT-4o, with a marginally positive $R^{2}$ (0.0347) and EV (0.0378), but still falls short compared to WELLA. These results highlight WELLA’s robustness and effectiveness in Scenario RO2.

\begin{table}[h]
\caption{ Performance Comparison of Models in Scenario RO3}\label{RO3-table}%
\begin{tabular}{p{4cm}p{2cm}p{2cm}p{2cm}p{2cm}}
\toprule
 & $R^{2} $  & RMSE  & MAE & EV \\
\midrule
GPT-4& -2.0091 & 31.7849 &19.6400 &-1.76788\\
GPT-4o&-4.6050 &43.3797 &34.4400 & -2.9852\\
Claude-3.5-Sonnet&-1.9544& 31.4948&18.4000 & -1.5861\\
WELLA&0.9628 &3.5327&1.9200 & 0.9666\\

\bottomrule
\end{tabular}
\end{table}

Table \ref{RO3-table} summarizes the performance of different models under Scenario RO3. WELLA demonstrates exceptional performance, achieving the highest $R^{2}$ (0.9628) and EV (0.9666), along with the lowest RMSE (3.5327) and MAE (1.9200), highlighting its strong predictive accuracy and ability to explain variance in the data. In stark contrast, GPT-4, GPT-4o, and Claude-3.5-Sonnet exhibit negative $R^{2}$  and EV values, coupled with significantly higher RMSE and MAE, indicating poor predictive capabilities. Among these models, GPT-4o shows the weakest performance with the lowest $R^{2}$  (-4.6050) and highest RMSE (43.3797) and MAE (34.4400). These results underscore WELLA’s superiority in handling complex predictions in scenario RO3.

\begin{table}[h]
\caption{ Performance Comparison of Models in Scenario CO}\label{CO-table}%
\begin{tabular}{p{4cm}p{2cm}p{2cm}p{2cm}p{2cm}}
\toprule
 & $R^{2} $  & RMSE  & MAE & EV \\
\midrule
GPT-4& -7.5568 & 41.8657 &31.9706 & -2.8206\\
GPT-4o&-4.029 &32.0973 &24.4706 & -2.0997\\
Claude-3.5-Sonnet& -2.3948&26.3701&19.2647 &  -0.7557\\
WELLA&0.4216 &10.8844&6.5882 & 0.4448\\

\bottomrule
\end{tabular}
\end{table}

Table \ref{CO-table} presents the performance comparison of models under Scenario CO. WELLA outperforms the other models, achieving the highest $R^{2}$ (0.4216) and EV (0.4448), along with the lowest RMSE (10.8844) and MAE (6.5882), indicating superior predictive accuracy and variance explanation. In contrast, GPT-4, GPT-4o, and Claude-3.5-Sonnet perform poorly, with negative $R^{2}$ and EV values, as well as significantly higher RMSE and MAE. Among them, GPT-4 has the weakest performance, with the lowest $R^{2}$ (-7.5568) and the highest RMSE (41.8657) and MAE (31.9706). These results clearly highlight WELLA’s robustness and reliability in Scenario CO.

\begin{table}[h]
\caption{ Performance Comparison of Models in Scenario SO}\label{SO-table}%
\begin{tabular}{p{4cm}p{2cm}p{2cm}p{2cm}p{2cm}}
\toprule
 & $R^{2} $  & RMSE  & MAE & EV \\
\midrule
GPT-4& -14.8660 & 41.2702 &30.1935 & -6.3737\\
GPT-4o& -9.6419 &33.7997 &24.0322 & -5.5086\\
Claude-3.5-Sonnet& -10.7456& 35.5092&26.5806& -4.1641\\
WELLA& 0.3822 &8.1439&4.5161 &0.4607\\

\bottomrule
\end{tabular}
\end{table}

Table \ref{SO-table} provides a comparison of model performance in Scenario SO. WELLA demonstrates outstanding performance, achieving the highest $R^{2} $ (0.3822) and EV (0.4607), alongside the lowest RMSE (8.1439) and MAE (4.5161), indicating its superior predictive accuracy and capability to explain data variability. In contrast, GPT-4, GPT-4o, and Claude-3.5-Sonnet exhibit significantly poorer results, with highly negative 
$R^{2} $ and EV values, as well as substantially higher RMSE and MAE. GPT-4 shows the weakest performance, with the lowest $R^{2} $  (-14.8660) and the highest RMSE (41.2702) and MAE (30.1935). These results emphasize WELLA’s robustness and effectiveness in Scenario SO.

\begin{table}[h]
\caption{ Performance Comparison of Models in Scenario ALL data}\label{ALL-table}%
\begin{tabular}{p{4cm}p{2cm}p{2cm}p{2cm}p{2cm}}
\toprule
 & $R^{2} $  & RMSE  & MAE & EV \\
\midrule
GPT-4& -14.8660 & 41.2702 &30.1935 & -6.3737\\
GPT-4o& -9.6419 &33.7997 &24.0322 & -5.5086\\
Claude-3.5-Sonnet& -10.7456& 35.5092&26.5806& -4.1641\\
WELLA& 0.3822 &8.1439&4.5161 &0.4607\\

\bottomrule
\end{tabular}
\end{table}
Table \ref{ALL-table} presents a performance comparison of models on the combined dataset. WELLA consistently outperforms all other models, achieving the highest 
$R^{2} $ (0.3822) and EV (0.4607), as well as the lowest RMSE (8.1439) and MAE (4.5161), indicating its superior predictive accuracy and ability to explain data variability. In contrast, GPT-4, GPT-4o, and Claude-3.5-Sonnet exhibit substantially poorer performance, with highly negative $R^{2}$ and EV values and significantly higher RMSE and MAE. Among them, GPT-4 shows the weakest results, with the lowest $R^{2} $ (-14.8660), the highest RMSE (41.2702), and the highest MAE (30.1935). These findings highlight WELLA’s robustness and effectiveness across the entire dataset.

% \begin{table}[h]
% \caption{ Scenario Settings}\label{data}%
% \begin{tabular}{p{3cm}p{3cm}p{3cm}p{3cm}}
% \toprule
% iter & good case  & bad case  & invalid  \\
% \midrule
%  1& 72 & 121 &221\\
% \bottomrule
% \end{tabular}
% \end{table}

% zero-shot
% 
% 第一轮迭代，一共414条测试数据中，被discriminator诊断结果完全一模一样的数据有72条，而错误的数据有96条，无效的数据有221条.
%其实这个是很难得得结果，因为对于回归问题，完全一摸一样是很困难的，常见的指标包括R2， MSE， 

% 但是，由于我们想提高准确率，因此设置了迭代更新muti-agent机制，实现对错误数据和无效data的update.

% 对比之下，gpt4、gpt4o、claude的结果完全一摸一样的结果都是0

% no special完全一摸一样的为4条,

% 在经过所提出的muti-agnet框架后

% 在discriminator的选择上，我们测试了claude3.5-sonnet 和 gpt4o，结果显示们，4o的效果更好，提供了更为清晰的案例分析及简化，而claude3.5-sonnet 无法正确的理解问题，会出现回答我所需要简化的问题，而不是反给我分析简化结果。

\subsection{Comparisons with Existing Methods}

The most widely used model for HRA synthetic data generation is the human unimodel for nuclear technology to enhance reliability (HUNTER) \cite{ulrich2022hunter}. It is a procedurally driven nuclear operator simulation developed to support dynamic HRA research.The underlying logic is based on the goals, operators, methods, and selection rules (GOMS) framework \cite{boring2016goms}. The simulation is designed to operate alongside a plant model, enabling dynamic performance context evaluations through the use of GOMS-HRA primitives to define time distributions and base human error probabilities for individual procedure steps. Despite its utility, GOMS has been criticized for being time-consuming and labor-intensive to model \cite{preece2002interaction}.

\begin{figure}[h]
\centering
\includegraphics[width=1.0\textwidth]{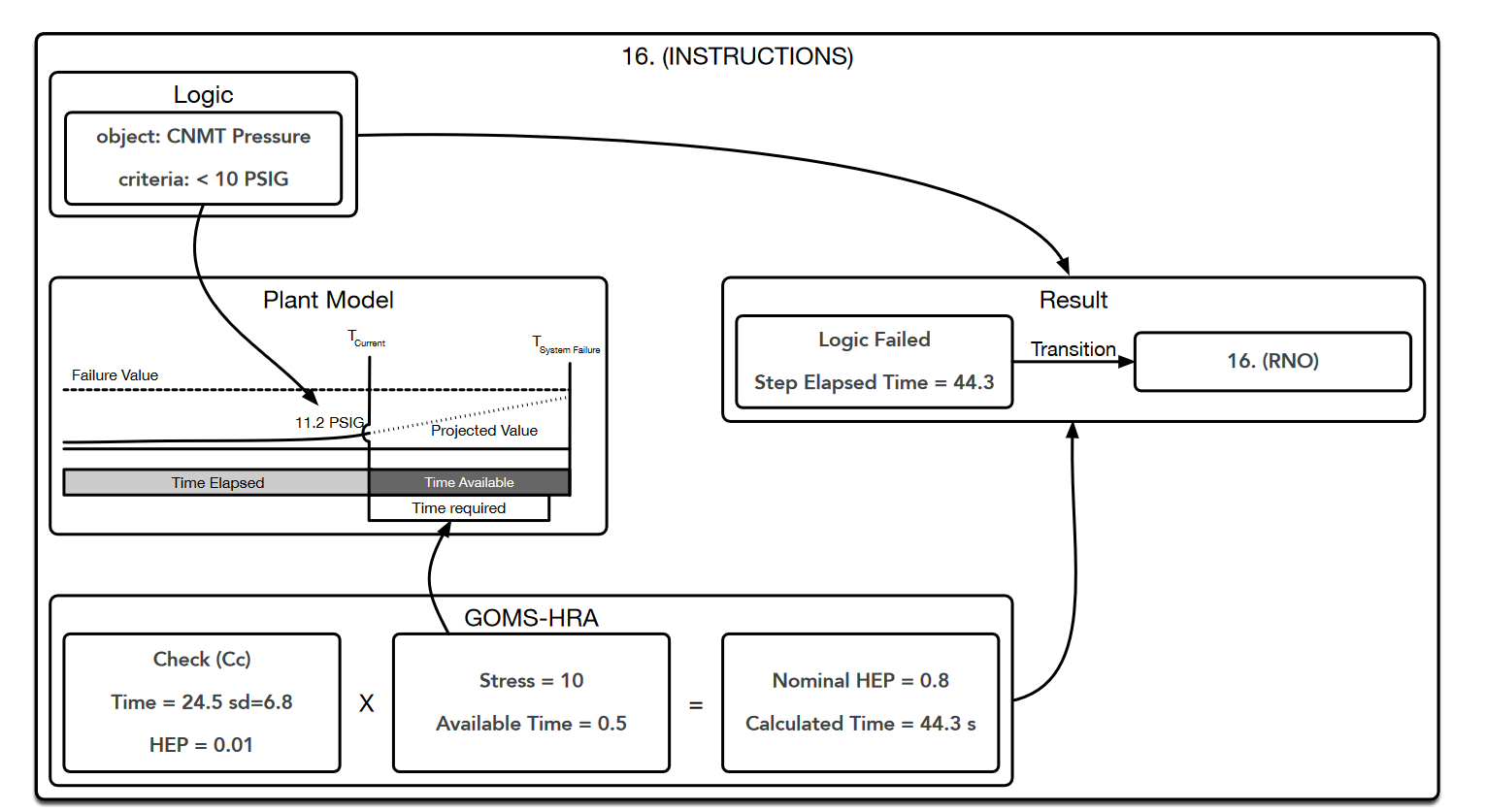}
\caption{Step Module Execution Example Using the Instruction Step \cite{ulrich2022hunter}. }\label{Hunter}
\end{figure}

Figure \ref{Hunter} illustrates a well-defined HUNTER process. The logic section first defines the logical conditions, specifying the objects to be checked and their associated criteria. The Plant Model analyzes factors such as pressure changes and time calculations, ultimately determining the actual task execution time. This execution time is then compared with the available time, and if the actual time exceeds the available time, the logic is considered to have failed, triggering the system to transition from the current step (16. Instructions) to the next operation (16. RNO). While this approach demonstrates a degree of innovation, it requires the input of these parameters each time, limiting its capacity for real-time processing. Additionally, the Monte Carlo simulation calculation demands a certain amount of time.

However, it can be observed that the human error judgment logic in HUNTER is relatively simplistic, as it merely compares time without considering the underlying factors contributing to human error. In contrast, our workload metric is more of a retrospective indicator, offering a deeper insight into human performance. Additionally, our approach integrates macro-cognitive theory, exploring human factors through the stages of detection, understanding, decision-making, action execution, and inter-team coordination. Furthermore, our method minimizes the need for excessive user input, unlike HUNTER, which requires inputs such as performance shaping factors (PSF), which rely on expert knowledge and subjective analysis. Table \ref{compare} provides a detailed comparison between our method, WELLA, and HUNTER.

\begin{table}[h]
\caption{Comparative Analysis between HUNTER and WELLA}\label{compare}%
\begin{tabular}{p{3cm}p{6cm}p{5cm}}
\toprule
  & HUNTER & WELLA \\
\midrule
 Input  &  (1) tasks, procedures, and steps mapped to specific plant parameters; (2) performance shaping factors with levels mapped to particular steps or the general equations for their calculation throughout the simulation; and (3) meta-simulation parameters, including the number of Monte Carlo runs and the result output format of the simulation parameters as well as the task and procedure objects. &(1) scenario (task) (2) A brief description of human job functions, such as "operating Reactor 1 and Reactor 2."  \\

 Output & Whether a human makes an error (evaluated from the perspective of execution time) & Workload and Situation Awareness (SA) through the lens of macro-cognitive theory \\
 Automation & It requires the integration of expert knowledge and manual input. & Fully automated \\
 Generalizability & Yes & No\\
 
\hline
\end{tabular}
\end{table}

\subsection{Scenario Analysis}\label{Scenario Analysis}

Next, we will test the performance of our model in substituting control room personnel across different scenarios. 

Given the scarcity of operational data in nuclear power plants, especially for high-temperature gas-cooled reactors (HTGR), this methodology offers a novel solution to bridge the data gap. By leveraging WELLA, the proposed approach can simulate realistic workload scenarios based on predefined protocols, thereby generating synthetic workload data for analysis and training purposes. This case study demonstrates how the methodology can be applied to create workload data in the absence of comprehensive real-world datasets, providing valuable insights into system behavior and operator performance in complex operational settings. The ability to automatically generate workload data also offers significant advantages for improving safety training, system design, and operational planning in nuclear power plants, where data availability is often limited.

\begin{figure}[H]
\centering
\includegraphics[width=1.0\textwidth]{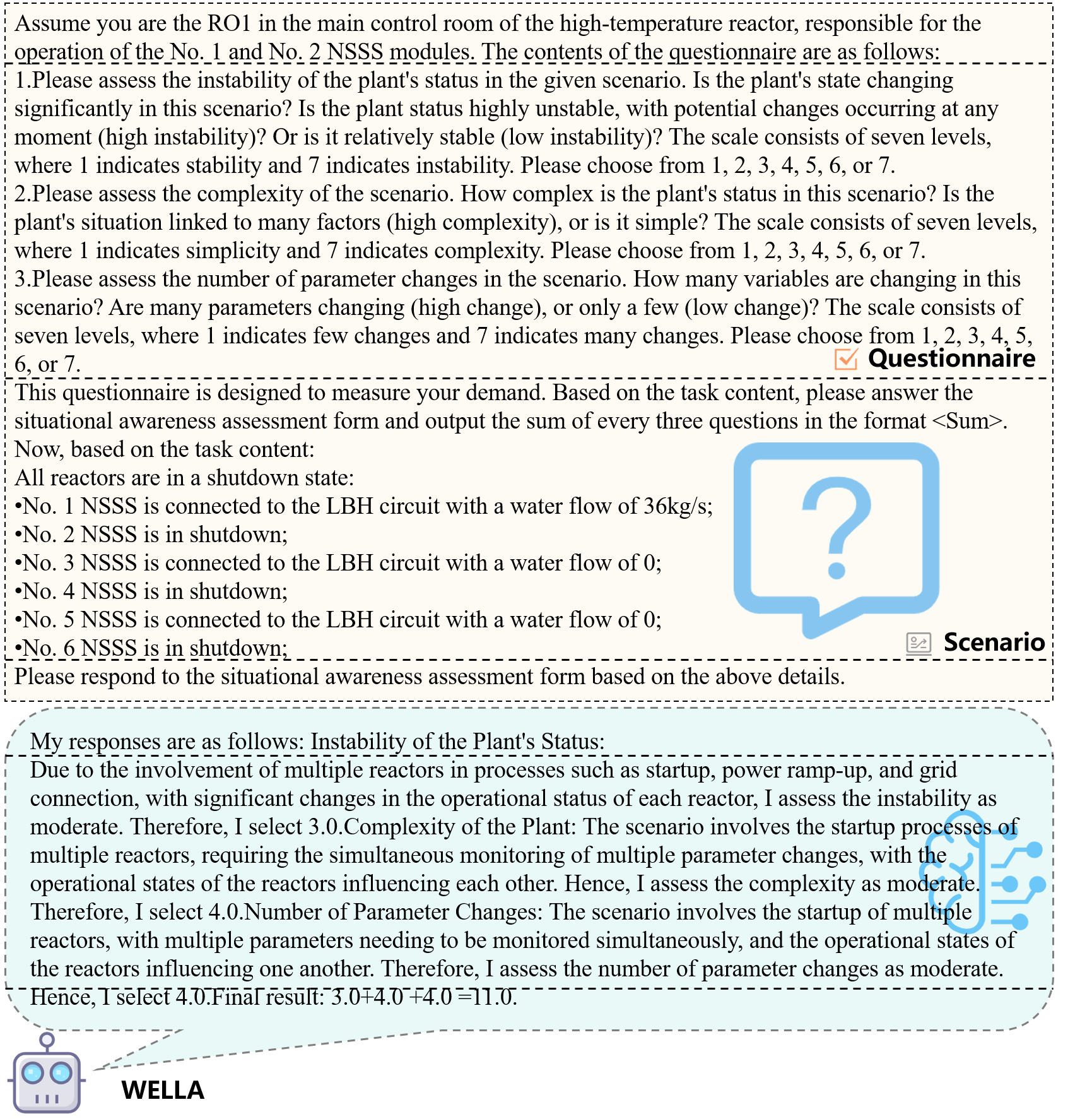}
\caption{The predictive analysis results of WELLA for RO1 in a specific scenario.. }\label{case}
\end{figure}

Figure \ref{case} demonstrates the application of the VELLA model for situational awareness assessment in the context of high-temperature reactor operations, specifically focusing on the No. 1 and No. 2 NSSS (Nuclear Steam Supply System) modules. The upper section of the figure represents the model's input, including a structured scenario detailing the operational states of six NSSS units (e.g., water flow rates and shutdown statuses) and a standardized questionnaire assessing three key factors on a 7-point Likert scale: instability of the plant’s status, complexity of the plant’s situation, and the number of parameter changes. The lower section illustrates the model’s output, encompassing a virtual cognitive journey that simulates the operator’s reasoning process and the final results of the questionnaire. In this example, the operator assessed the scenario with scores of 3 (instability), 4 (complexity), and 4 (parameter changes), yielding a total situational demand score of 11. By integrating scenario analysis, cognitive reasoning, and standardized evaluation, the VELLA model provides a structured and transparent framework for assessing situational awareness in complex operational environments.

\section{Conclusion and Discussion}

This study presents a pioneering approach to workload estimation in human reliability analysis (HRA) by leveraging fine-tuned large language models (LLMs) in real-world collaborative scenarios. Our method addresses the limitations of traditional data collection techniques by automating the gathering of HRA data and capturing dynamic, real-time cognitive load across different operator roles. Through training LLMs on operational data from high-temperature gas-cooled reactors (HTGRs), we are able to simulate human behavior with greater accuracy and adaptability. The results show that WELLA (Workload Estimation with LLMs and Agents) provides superior performance compared to existing commercial LLM-based methods, offering a more flexible, scalable, and efficient solution for estimating operator workload. This advancement has the potential to enhance safety and decision-making in complex, multi-agent systems and can be extended to other domains requiring dynamic workload assessments. Future work could explore the integration of virtual digital human technologies to expand simulation through 3D scene modeling, thereby constructing a virtual world \cite{chen2024mixedgaussianavatar} \cite{chen2023diffusiontalker}. While the current approach is scenario-driven, it has not yet fully incorporated the specific characteristics of operators and nuclear power plants. Subsequent research could consider integrating these two aspects to further enhance and refine the study.

% 1. 针对的是高温气冷堆，一人带两堆
% 2. 使用的是真实操作员的数据
% 3。 针对的是多人协作场景
% 4. 是一种动态的方法，自动化方法
% 5. 可用于根据不同的场景来产生workload数据
% 6. 是一种情景驱动的方法，不同的情景有不同的回答
% 7. 是一种real-time ，可以预警
% 8. 提出了做这种实验的新范式。

\appendix

 \bibliographystyle{elsarticle-num} 
 \bibliography{cas-refs}

\begin{thebibliography}{10}
\expandafter\ifx\csname url\endcsname\relax
  \def\url#1{\texttt{#1}}\fi
\expandafter\ifx\csname urlprefix\endcsname\relax\def\urlprefix{URL }\fi
\expandafter\ifx\csname href\endcsname\relax
  \def\href#1#2{#2} \def\path#1{#1}\fi

\bibitem{xiao2024krail}
X.~Xiao, P.~Chen, B.~Qi, H.~Zhao, J.~Liang, J.~Tong, H.~Wang, Krail: A knowledge-driven framework for base human reliability analysis integrating idheas and large language models, arXiv preprint arXiv:2412.18627 (2024).

\bibitem{yu2024human}
Y.~Yu, S.~Wu, Y.~Fu, X.~Liu, Q.~Zeng, H.~Ding, Y.~Pan, Y.~Wu, H.~Guo, Y.~Yang, Human reliability analysis of offshore high integrity pressure protection system based on improved cream and hcr integration method, Ocean Engineering 307 (2024) 118153.

\bibitem{gertman2005spar}
D.~Gertman, H.~Blackman, J.~Marble, J.~Byers, C.~Smith, et~al., The spar-h human reliability analysis method, US Nuclear Regulatory Commission 230~(4) (2005) 35.

\bibitem{hollnagel1998cognitive}
E.~Hollnagel, Cognitive reliability and error analysis method (CREAM), Elsevier, 1998.

\bibitem{xing2020integrated}
J.~Xing, Y.~Chang, J.~DeJesus, Integrated human event analysis system for event and condition assessment (idheas-eca), US Nuclear Regulatory Commission, Washington, DC (2020).

\bibitem{xing2021draft}
J.~Xing, Y.~CHANG, J.~SEGARRA, Draft—integrated human event analysis system for human reliability data (idheas-data), RIL-2021-XX (2021).

\bibitem{jung2020hurex}
W.~Jung, J.~Park, Y.~Kim, S.~Y. Choi, S.~Kim, Hurex--a framework of hra data collection from simulators in nuclear power plants, Reliability Engineering \& System Safety 194 (2020) 106235.

\bibitem{chang2014sacada}
Y.~J. Chang, D.~Bley, L.~Criscione, B.~Kirwan, A.~Mosleh, T.~Madary, R.~Nowell, R.~Richards, E.~M. Roth, S.~Sieben, et~al., The sacada database for human reliability and human performance, Reliability Engineering \& System Safety 125 (2014) 117--133.

\bibitem{boring2016goms}
R.~Boring, M.~Rasmussen, Goms-hra: A method for treating subtasks in dynamic human reliability analysis, in: Proceedings of the 2016 European Safety and Reliability Conference, 2016, pp. 956--963.

\bibitem{boring2023procedure}
R.~L. Boring, T.~A. Ulrich, R.~Lew, The procedure performance predictor (p3): Application of the hunter dynamic human reliability analysis software to inform the development of new procedures (2023).

\bibitem{xiao2024emergency}
X.~Xiao, J.~Liang, J.~Tong, H.~Wang, Emergency decision support techniques for nuclear power plants: Current state, challenges, and future trends, Energies 17~(10) (2024) 2439.

\bibitem{boring2007measure}
R.~L. Boring, C.~D. Griffith, J.~C. Joe, The measure of human error: Direct and indirect performance shaping factors, in: 2007 IEEE 8th Human Factors and Power Plants and HPRCT 13th Annual Meeting, IEEE, 2007, pp. 170--176.

\bibitem{chamon2013iceberg}
M.~Chamon, E.~Kaplan, The iceberg theory of campaign contributions: Political threats and interest group behavior, American Economic Journal: Economic Policy (2013) 1--31.

\bibitem{wright2000towards}
L.~Wright, Towards an empirical test of the iceberg model, Hum. Decis. Mak. Man. Control. EAM2000 EUR 19599 (2000) 145--152.

\bibitem{domurath2015stress}
J.~Domurath, M.~Saphiannikova, J.~F{\'e}rec, G.~Ausias, G.~Heinrich, Stress and strain amplification in a dilute suspension of spherical particles based on a bird--carreau model, Journal of Non-Newtonian Fluid Mechanics 221 (2015) 95--102.

\bibitem{kantowitz2017human}
B.~H. Kantowitz, P.~A. Casper, Human workload in aviation, in: Human error in aviation, Routledge, 2017, pp. 123--153.

\bibitem{jou2009evaluation}
Y.-T. Jou, T.-C. Yenn, C.~J. Lin, C.-W. Yang, C.-C. Chiang, Evaluation of operators’ mental workload of human--system interface automation in the advanced nuclear power plants, Nuclear Engineering and Design 239~(11) (2009) 2537--2542.

\bibitem{bagheri2020eeg}
M.~Bagheri, S.~D. Power, Eeg-based detection of mental workload level and stress: the effect of variation in each state on classification of the other, Journal of Neural Engineering 17~(5) (2020) 056015.

\bibitem{brouwer2012estimating}
A.-M. Brouwer, M.~A. Hogervorst, J.~B. Van~Erp, T.~Heffelaar, P.~H. Zimmerman, R.~Oostenveld, Estimating workload using eeg spectral power and erps in the n-back task, Journal of neural engineering 9~(4) (2012) 045008.

\bibitem{hassan2024eeg}
J.~Hassan, M.~S. Reza, S.~U. Ahmed, N.~H. Anik, M.~O. Khan, Eeg workload estimation and classification: a systematic review, Journal of Neural Engineering (2024).

\bibitem{delliaux2019mental}
S.~Delliaux, A.~Delaforge, J.-C. Deharo, G.~Chaumet, Mental workload alters heart rate variability, lowering non-linear dynamics, Frontiers in physiology 10 (2019) 565.

\bibitem{marquart2015review}
G.~Marquart, C.~Cabrall, J.~De~Winter, Review of eye-related measures of drivers’ mental workload, Procedia Manufacturing 3 (2015) 2854--2861.

\bibitem{planke2021online}
L.~J. Planke, A.~Gardi, R.~Sabatini, T.~Kistan, N.~Ezer, Online multimodal inference of mental workload for cognitive human machine systems, Computers 10~(6) (2021) 81.

\bibitem{xing2018driver}
Y.~Xing, C.~Lv, D.~Cao, H.~Wang, Y.~Zhao, Driver workload estimation using a novel hybrid method of error reduction ratio causality and support vector machine, Measurement 114 (2018) 390--397.

\bibitem{schvaneveldt1998modeling}
R.~W. Schvaneveldt, G.~B. Reid, R.~Gomez, S.~Rice, Modeling mental workload, Cognitive Technology 3~(1) (1998) 19--31.

\bibitem{ritter2019act}
F.~E. Ritter, F.~Tehranchi, J.~D. Oury, Act-r: A cognitive architecture for modeling cognition, Wiley Interdisciplinary Reviews: Cognitive Science 10~(3) (2019) e1488.

\bibitem{liu2006queueing}
Y.~Liu, R.~Feyen, O.~Tsimhoni, Queueing network-model human processor (qn-mhp) a computational architecture for multitask performance in human-machine systems, ACM Transactions on Computer-Human Interaction (TOCHI) 13~(1) (2006) 37--70.

\bibitem{paxion2014mental}
J.~Paxion, E.~Galy, C.~Berthelon, Mental workload and driving, Frontiers in psychology 5 (2014) 1344.

\bibitem{longo2018experienced}
L.~Longo, Experienced mental workload, perception of usability, their interaction and impact on task performance, PloS one 13~(8) (2018) e0199661.

\bibitem{dehais2020neuroergonomics}
F.~Dehais, A.~Lafont, R.~Roy, S.~Fairclough, A neuroergonomics approach to mental workload, engagement and human performance, Frontiers in neuroscience 14 (2020) 268.

\bibitem{wang2024adapting}
K.~Wang, Y.~Lu, M.~Santacroce, Y.~Gong, C.~Zhang, et~al., Adapting llm agents with universal feedback in communication, in: ICML 2024 Workshop on Foundation Models in the Wild, 2024.

\bibitem{xie2024can}
C.~Xie, C.~Chen, F.~Jia, Z.~Ye, K.~Shu, A.~Bibi, Z.~Hu, P.~Torr, B.~Ghanem, G.~Li, Can large language model agents simulate human trust behaviors?, arXiv preprint arXiv:2402.04559 (2024).

\bibitem{sreedhar2025simulating}
K.~Sreedhar, L.~Chilton, Simulating strategic reasoning: Comparing the ability of single llms and multi-agent systems to replicate human behavior (2025).

\bibitem{wu2024shall}
Z.~Wu, R.~Peng, S.~Zheng, Q.~Liu, X.~Han, B.~Kwon, M.~Onizuka, S.~Tang, C.~Xiao, Shall we team up: Exploring spontaneous cooperation of competing llm agents, in: Findings of the Association for Computational Linguistics: EMNLP 2024, 2024, pp. 5163--5186.

\bibitem{tjuatja2024llms}
L.~Tjuatja, V.~Chen, T.~Wu, A.~Talwalkwar, G.~Neubig, Do llms exhibit human-like response biases? a case study in survey design, Transactions of the Association for Computational Linguistics 12 (2024) 1011--1026.

\bibitem{xiao2024text}
X.~Xiao, S.~Liu, Z.~Zuo, P.~Chen, B.~Qi, J.~Liang, J.~Tong, A text intelligence-based approach for automatic generation of fault trees in nuclear power plants, in: International Conference on Nuclear Engineering, Vol. 88308, American Society of Mechanical Engineers, 2024, p. V010T12A004.

\bibitem{zhang2006design}
Z.~Zhang, Z.~Wu, Y.~Sun, F.~Li, Design aspects of the chinese modular high-temperature gas-cooled reactor htr-pm, Nuclear Engineering and Design 236~(5-6) (2006) 485--490.

\bibitem{hart1986nasa}
S.~G. Hart, Nasa task load index (tlx) (1986).

\bibitem{taylor2017situational}
R.~M. Taylor, Situational awareness rating technique (sart): The development of a tool for aircrew systems design, in: Situational awareness, Routledge, 2017, pp. 111--128.

\bibitem{hart2006nasa}
S.~G. Hart, Nasa-task load index (nasa-tlx); 20 years later, in: Proceedings of the human factors and ergonomics society annual meeting, Vol.~50, Sage publications Sage CA: Los Angeles, CA, 2006, pp. 904--908.

\bibitem{strydom2011cognitive}
P.~Strydom, The cognitive and metacognitive dimensions of social and political theory, in: Routledge International Handbook of Contemporary Social and Political Theory, Routledge, 2011, pp. 328--338.

\bibitem{dong2023abilities}
G.~Dong, H.~Yuan, K.~Lu, C.~Li, M.~Xue, D.~Liu, W.~Wang, Z.~Yuan, C.~Zhou, J.~Zhou, How abilities in large language models are affected by supervised fine-tuning data composition, arXiv preprint arXiv:2310.05492 (2023).

\bibitem{yang2024qwen2}
A.~Yang, B.~Yang, B.~Zhang, B.~Hui, B.~Zheng, B.~Yu, C.~Li, D.~Liu, F.~Huang, H.~Wei, et~al., Qwen2. 5 technical report, arXiv preprint arXiv:2412.15115 (2024).

\bibitem{zheng2024llamafactory}
Y.~Zheng, R.~Zhang, J.~Zhang, Y.~Ye, Z.~Luo, Z.~Feng, Y.~Ma, Llamafactory: Unified efficient fine-tuning of 100+ language models, arXiv preprint arXiv:2403.13372 (2024).

\bibitem{liu2023hierarchical}
X.~Liu, J.~Zhang, H.~Zhang, F.~Xue, Y.~You, Hierarchical dialogue understanding with special tokens and turn-level attention, arXiv preprint arXiv:2305.00262 (2023).

\bibitem{achiam2023gpt}
J.~Achiam, S.~Adler, S.~Agarwal, L.~Ahmad, I.~Akkaya, F.~L. Aleman, D.~Almeida, J.~Altenschmidt, S.~Altman, S.~Anadkat, et~al., Gpt-4 technical report, arXiv preprint arXiv:2303.08774 (2023).

\bibitem{shahriar2024putting}
S.~Shahriar, B.~D. Lund, N.~R. Mannuru, M.~A. Arshad, K.~Hayawi, R.~V.~K. Bevara, A.~Mannuru, L.~Batool, Putting gpt-4o to the sword: A comprehensive evaluation of language, vision, speech, and multimodal proficiency, Applied Sciences 14~(17) (2024) 7782.

\bibitem{xiao2024hybrid}
X.~Xiao, P.~Chen, A hybrid real-time framework for efficient fussell-vesely importance evaluation using virtual fault trees and graph neural networks, arXiv preprint arXiv:2412.10484 (2024).

\bibitem{ulrich2022hunter}
T.~Ulrich, R.~Boring, J.~Park, Y.~Heo, J.~Ahn, The hunter dynamic human reliability analysis tool: Procedurally driven operator simulation, in: 16th International Conference on Probabilistic Safety Assessment and Management, PSAM 2022, 2022.

\bibitem{preece2002interaction}
J.~Preece, Interaction issues for special applications, in: The Human-Computer Interaction Handbook, CRC Press, 2002, pp. 555--784.

\bibitem{chen2024mixedgaussianavatar}
P.~Chen, X.~Wei, Q.~Wuwu, X.~Wang, X.~Xiao, M.~Lu, Mixedgaussianavatar: Realistically and geometrically accurate head avatar via mixed 2d-3d gaussian splatting, arXiv preprint arXiv:2412.04955 (2024).

\bibitem{chen2023diffusiontalker}
P.~Chen, X.~Wei, M.~Lu, Y.~Zhu, N.~Yao, X.~Xiao, H.~Chen, Diffusiontalker: Personalization and acceleration for speech-driven 3d face diffuser, arXiv preprint arXiv:2311.16565 (2023).

\end{thebibliography}





% \end{thebibliography}
\end{document}